\documentclass{article}

\usepackage{microtype}
\usepackage{graphicx}
\usepackage{subfigure}
\usepackage{booktabs} 

\usepackage{hyperref}

\usepackage[accepted]{icml2023}

\usepackage{amsmath}
\usepackage{amssymb}
\usepackage{mathtools}
\usepackage{amsthm}

\usepackage[capitalize,noabbrev]{cleveref}

\theoremstyle{plain}

\theoremstyle{definition}

\theoremstyle{remark}

\usepackage[textsize=tiny]{todonotes}

\usepackage{paralist}
\usepackage{siunitx}
\usepackage{adjustbox}
\usepackage{multirow}

\usepackage{pifont}
\newcommand{\cmark}{\text{\ding{51}}}
\newcommand{\xmark}{\text{\ding{55}}}
\DeclareMathOperator*{\E}{\mathbb{E}}

\usepackage{expl3}
\ExplSyntaxOn
\newcommand\latinabbrev[1]{
  \peek_meaning:NTF . {
    #1\@}%
  { \peek_catcode:NTF a {
      #1.\@ }%
    {#1.\@}}}
\ExplSyntaxOff

\def\ie{\latinabbrev{i.e}}

\icmltitlerunning{Long-Tailed Recognition by Mutual Information Maximization between Latent Features and Ground-Truth Labels}

\begin{document}

\twocolumn[
\icmltitle{Long-Tailed Recognition by Mutual Information Maximization \\
           between Latent Features and Ground-Truth Labels}



\icmlsetsymbol{equal}{*}

\begin{icmlauthorlist}
\icmlauthor{Min-Kook Suh}{snu}
\icmlauthor{Seung-Woo Seo}{snu}
\end{icmlauthorlist}

\icmlaffiliation{snu}{Department of Electrical and Computer Engineering, Seoul National University, Seoul, South Korea}
\icmlcorrespondingauthor{Min-Kook Suh}{bluecdm@snu.ac.kr}

\icmlkeywords{Long-tailed recognition, Mutual information maximization, Contrastive learning, Image classification, Semantic segmentation}

\vskip 0.3in
]



\printAffiliationsAndNotice{}  

\begin{abstract}

Although contrastive learning methods have shown prevailing performance on a variety of representation learning tasks, they encounter difficulty when the training dataset is long-tailed. Many researchers have combined contrastive learning and a logit adjustment technique to address this problem, but the combinations are done ad-hoc and a theoretical background has not yet been provided. The goal of this paper is to provide the background and further improve the performance. First, we show that the fundamental reason contrastive learning methods struggle with long-tailed tasks is that they try to maximize the mutual information between latent features and input data. As ground-truth labels are not considered in the maximization, they are not able to address imbalances between classes. Rather, we interpret the long-tailed recognition task as a mutual information maximization between latent features and ground-truth labels. This approach integrates contrastive learning and logit adjustment seamlessly to derive a loss function that shows state-of-the-art performance on long-tailed recognition benchmarks. It also demonstrates its efficacy in image segmentation tasks, verifying its versatility beyond image classification. Code is available at \url{https://github.com/bluecdm/Long-tailed-recognition}.

\end{abstract}
\section{Introduction}
\label{sec:intro}

A supervised classification task has been an active research topic for decades. However, its performance is still unsatisfactory when the dataset shows a long-tailed distribution, where a few classes, \ie, head classes, contain a major number of samples while the remaining classes, \ie, tail classes, have only a small number of samples. The most straightforward remedy to this problem is to rebalance the training dataset through weighted sampling. However, it is a suboptimal strategy that may be detrimental to the accuracy of the head classes~\cite{wang2017learning}.

\begin{figure*}[t]
    \centering
    \includegraphics[width=0.91\textwidth]{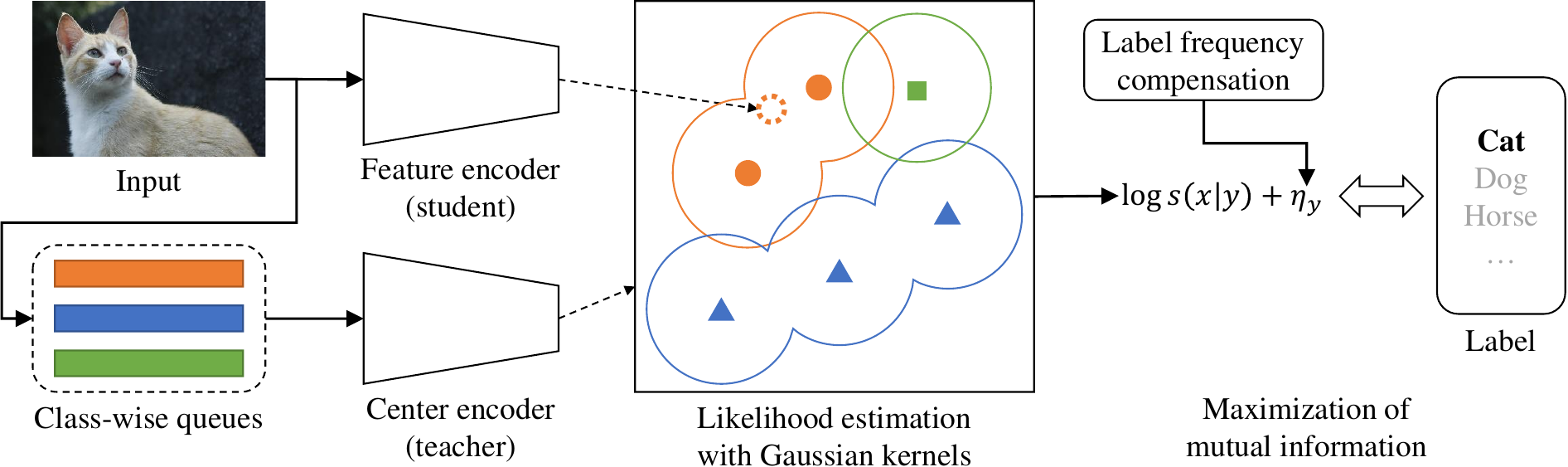}
    \caption{We address the long-tailed recognition by maximizing the mutual information of latent features, $x$, and ground-truth labels, $y$. We solve the maximization problem by dividing it into two terms: an unnormalized likelihood term, $s(x|y)$, and a logit-adjustment term to compensate for label frequency, $\eta_y$. $\eta_y$ is a value that depends on the frequency of the label, and $s(x|y)$ is estimated using a neural network. The proposed loss is achieved by estimating $s(x|y)$ with Gaussian kernels using latent features of other samples.}
    \label{fig:overview}
\end{figure*}

Recently, contrastive learning~\cite{oord2018representation,chen2020simple,khosla2020supervised} is widely used in representation learning and showing state-of-the-art performance on various tasks. The contrastive learning framework learns representations by pushing latent features from the same sample closer and separating them from different samples. However, its performance degrades on long-tailed datasets as the samples are imbalanced~\cite{cui2021parametric}. Therefore, several works have been conducted to adapt it to the long-tailed recognition tasks~\cite{cui2021parametric,zhu2022balanced} by combining it with logit adjustment techniques~\cite{ren2020balanced,menon2021long}, which is another method to solve the long-tailed recognition task that modulates the prediction score of networks based on the appearance frequency of classes. Although previous methods empirically show that combining contrastive learning and logit adjustment boosts performance, they do not provide the theoretical meaning of the combination.

In this paper, we describe the theoretical meaning of the combination and provide an improved method for combining contrastive learning and logit adjustment. We find that the performance of previous contrastive learning methods degrade on long-tailed datasets because they aim to maximize the mutual information (MI) between the latent features and input data. These approaches do not consider the imbalance of label distribution, as the ground-truth label is not involved in the maximization. Instead, we propose maximizing the MI between latent features and ground-truth labels, allowing the consideration of label distribution.

By replacing the input data term with the ground-truth label term, we derive a general loss function that encompasses various other methods. The derived loss function includes a likelihood of latent feature term and a prior of classes term, and different ways of modeling these terms lead to different previous methods. For example, the softmax cross-entropy loss for a balanced dataset and the logit adjustment term for an imbalanced dataset~\cite{ren2020balanced} can be derived by modeling the likelihood as an isotropic Gaussian. Supervised contrastive loss~\cite{khosla2020supervised} can be derived under the assumption of a balanced training dataset by estimating the likelihood using a sampling-based kernel density estimation with Gaussian kernels.

By removing the assumption of a balanced dataset, we derive the proposed loss function that seamlessly integrates the contrastive learning and logit adjustment. Since the kernel density estimation results in a Gaussian mixture likelihood, we refer to the loss as Gaussian Mixture Likelihood (GML) loss. We also provide an efficient method for modeling the Gaussian mixture likelihood, as shown in Fig.~\ref{fig:overview}. We use contrast samples to estimate the likelihood. Similar to Momentum Contrast (MoCo)~\cite{he2020momentum}, we use queues to store contrast samples. However, because a long-tailed dataset is being handled, tail classes do not have sufficient samples to create the Gaussian mixture. To resolve this problem, we use multiple class-wise queues rather than a single queue of MoCo. However, in the case of tail classes, the update frequencies of contrast samples are significantly longer. As a result, old samples of tail classes' queues are generated by highly outdated encoders. Therefore, we use class-wise queues along with a teacher--student strategy. Unlike a momentum encoder used in MoCo, we use a pre-trained teacher to generate contrast samples.

We evaluate the proposed method on various long-tailed recognition datasets: ImageNet-LT~\cite{liu2019large}, iNaturalist 2018~\cite{van2018inaturalist}, and CIFAR-LT~\cite{cui2019class}; the proposed method surpasses all previous long-tailed recognition methods. In addition, as the proposed method is related to supervised contrastive learning and knowledge distillation, we compare our method with them on both balanced and imbalanced datasets. Unsurprisingly, our method shows superior performance to them on imbalanced datasets, exhibiting comparable performance on balanced datasets. Finally, we evaluate our method on a commonly used semantic segmentation dataset ADE20K~\cite{zhou2017scene}, which also contains imbalanced pixel labels. Simply replacing the cross-entropy loss with the proposed loss also boosts the performance, indicating the proposed framework can be extended beyond a classification task.

Our main contributions are summarized as follows.
\begin{compactitem}
    \item We show that the fundamental limitation of contrastive learning methods on long-tailed tasks comes from directly maximizing the MI between latent features and input data. Instead, we propose to tackle the long-tailed recognition by MI maximization between latent features and ground-truth labels.
    \item While previous methods have combined contrastive learning and the logit adjustment without investigating a theoretical background, we find that contrastive learning implies a Gaussian mixture likelihood and the logit adjustment is derived from the prior of classes.
    \item We propose an efficient way to model the Gaussian mixture likelihood using a teacher--student framework that demonstrates its superiority in various long-tailed recognition tasks.
\end{compactitem}
\section{Related Work}
\subsection{Long-Tailed Recognition}
\textbf{Rebalancing Datasets.} Since the most straightforward approach to the long-tailed recognition problem is rebalancing the dataset during training, several early works~\cite{chawla2002smote,japkowicz2002class,drummond2003c4,han2005borderline,he2009learning,mikolov2013distributed} have focused on rebalancing approaches. 
However, \citet{byrd2019effect} found that the effect of rebalancing diminishes on overparameterized deep neural networks given sufficiently long training epochs.

\textbf{Normalized Classifier.} Networks trained on long-tailed datasets tend to have biased classifier weights~\cite{alshammari2022long}; this problem can be alleviated by normalizing the weights of the classification layer. \citet{kang2020decoupling} proposed to normalize the weights by decoupling the representation learning and classification. They first trained the network jointly using an instance-based sampling strategy. Then, they retrained the classifier only using a class-balanced sampling strategy.
\citet{gidaris2018dynamic} proposed the use of a cosine similarity classifier instead of a dot-product classifier. It bypasses the biased weights problem by only considering the relative angle. Accordingly, we adopt the cosine similarity as the classifier of our network.

\textbf{Logit Adjustment.} Another approach to the long-tailed recognition problem is to modulate the logit values. Interestingly, \citet{ren2020balanced} and \citet{menon2021long} derived similar results using different approaches. \citet{ren2020balanced} showed that the softmax loss is a biased estimator and proposed a Balanced Softmax loss; \citet{menon2021long} proposed a post-hoc logit adjustment and a logit-adjusted softmax cross-entropy loss. Both works show that adding a logit-adjustment term proportional to the logarithm of label frequency is essential to the long-tailed recognition. In accordance with their results, many studies~\cite{cui2021parametric,feng2021exploring,hong2021disentangling,zhu2022balanced} have included logit adjustment as a part of their methods.

\subsection{Contrastive Learning}
\citet{cui2021parametric} found that the performance of supervised contrastive loss~\cite{khosla2020supervised} significantly degrades when it is applied to a long-tailed dataset. Therefore, they proposed Parametric Contrastive learning (PaCo), which rebalances the contrast samples by adding parametric class-wise learnable centers in the samples. To integrate the logit-adjustment technique into their method, the authors added the adjustment term to the center learning. Further, \citet{zhu2022balanced} proposed Balanced Contrastive Learning (BCL), which utilizes the number of contrasts of each class in a mini-batch to balance the gradient contribution of classes. To integrate the logit-adjustment technique, they used a weighted sum of the logit-adjustment loss and their loss.
\section{Proposed Method}
In this section, we describe our approach to handle long-tailed recognition based on MI. Because the MI is intractable, we use InfoNCE loss~\cite{oord2018representation} to maximize its lower bound. We adopt the notations of \citet{poole2019variational} to express InfoNCE loss.
\begin{equation} \label{eq:infonce}
    I(X;Y) \ge \E\left[ \frac{1}{K}\sum^K_{i=1} \log \frac{\exp f(x_i,y_i)} {\frac{1}{K} \sum^K_{j=1}\exp f(x_i,y_j)} \right]
\end{equation}
where the expectation is taken over subsets of a dataset and $K$ denotes the size of the subset. Equality holds when $f(x,y)=\log p(x|y) + c(x)$ and $K\rightarrow \infty$ where $c(x)$ is a function that only depends on $x$.

\subsection{Contrastive Learning as MI Maximization between Latent Features and Input Data}
From Eq.~\ref{eq:infonce}, we can recover the loss functions of contrastive learning methods~\cite{chen2020big,he2020momentum} by substituting $x_i$ with the query feature, $f_q(t_q(x_i),w)$, and $y_j$ with the input image, $x_j$; it shows that contrastive learning methods maximize the MI between latent features and input data. Detailed proof is shown in Appendix~\ref{sec:detailed_contrastive_formulation}.

Since ground-truth label term is not included in the maximization, they are prone to the imbalance of label frequency. Supervised contrastive learning~\cite{khosla2020supervised} tries to modify the loss function to include the ground-truth label term, but it is still based on the MI maximization between latent features and input data.

\subsection{Long-Tailed Recognition by MI Maximization between Latent Features and Ground-Truth Labels} \label{sec:proposed_formulation}
To enable the label frequency considered in the MI maximization process, we formulate the long-tailed recognition problem as a maximum MI problem of latent features and ground-truth labels. However, replacing the input data term with the ground-truth label term results in a loss function that is not contrastive learning. In this section, we describe procedures to recover the contrastive loss.

\textbf{Logit Adjustment.} 
First, we substitute $X$ in Eq.~\ref{eq:infonce} with the set of latent features and $Y$ with the set of ground-truth labels to represent the MI maximization between latent features and ground-truth labels.
\begin{align}
    & \E_{i} \log \frac{\exp f(x_i,y_i)}{\E_{j} \exp f(x_i,y_j)}
    =\E_{i} \log \frac{\exp f(x_i,y_i)}{\sum_{c\in C} \exp f(x_i,c)p(c)} \nonumber \\
    & = \E_{i} \log \frac{\exp (f(x_i,y_i) + \eta_{y_i})}{\sum_{c\in C} \exp (f(x_i,c) + \eta_c)} - \eta_{y_i} \label{eq:logit_compensate}
\end{align}
We realize the above term by increasing $K$ to the size of the entire training dataset. Here, $x_i$ and $y_i$ denote the latent feature and ground-truth label of the $i$-th sample, respectively. $C$ denotes the set of all classes and $\eta_c=\log p(c)$ denotes the logit-adjustment term, which is the logarithm of the appearance frequency of a class.

Eq.~\ref{eq:logit_compensate} is a general template and various previous methods can be derived by different modeling of the likelihood and prior. For simplicity, we define $s(x|y)=\exp f(x,y)$ and use $s$ to denote the unnormalized likelihood of latent features. As an example, we derive a softmax cross-entropy loss with the logit-adjustment term by defining $s(x|y)=\exp (w_y\cdot x+b_y)$, a dot-product classifier.

\textbf{Gaussian Mixture Likelihood Loss.} The inequality on Eq.~\ref{eq:infonce} becomes tighter as $f$ approaches $\log p(x|y)$, hence, we choose to estimate $s$ using sampling-based kernel density estimation with Gaussian kernels. This estimation leads to a Gaussian mixture likelihood, where the centers of the Gaussian mixtures are the contrast samples.
\begin{equation} \label{eq:gaussian_likelihood}
    s(x|y)=\exp f(x,y)=\frac{1}{\lVert Z_y \rVert}\sum_{z_y \in Z_y} \exp (z_x \cdot z_y / \tau_g)
\end{equation}
where $z_x$ denotes the L2 normalized query feature of $x$, $Z_y$ denotes a set of L2 normalized contrast features of class $y$, and $\tau_g$ denotes a temperature parameter for GML loss, which is quadratically proportional to the variance of the Gaussian mixture. The subscript $i$ of $x_i$ and $y_i$ is omitted for simplicity. Gaussian mixture is represented using dot product instead of a quadratic term to maintain consistency with previous contrastive losses. It does not modify the meaning of Gaussian mixture, as L2 normalization is applied to both the query and contrast features. Specifically,
\begin{equation}
    -\lVert z_x - z_y \rVert^2_2 / (2\tau_g) = z_x \cdot z_y / \tau_g - 1 / \tau_g
\end{equation}
and the last constant term cancels out in the following equation.

Note that we do not estimate the centers of Gaussian mixtures, but simply use the latent features of the contrast encoder as centers. Therefore, the training burden is not increased by this procedure and remains almost the same as that of previous contrastive learning methods.

By modeling $s$ using the Gaussian mixture, we derive the GML loss which seamlessly integrates the contrastive learning and the logit compensation.
Substituting $f$ of Eq.~\ref{eq:logit_compensate} using Eq.~\ref{eq:gaussian_likelihood} derives the proposed loss $L_{GML}$.
\begin{align}
    &-L_{GML}=\\
    &\log\frac{\exp\left[\log\frac{1}{\lVert Z_y \rVert}\sum_{z_y \in Z_y} \exp(z_x\cdot z_y/\tau_g)+\eta_y \right]}
    {\sum_{c\in C}\exp\left[\log\frac{1}{\lVert Z_c \rVert}\sum_{z_c \in Z_c} \exp(z_x\cdot z_c/\tau_g)+\eta_c \right]} \nonumber
\end{align}

\subsection{Training with GML Loss}~\label{sec:training_gml}
\textbf{Class-wise Queues for Contrast Samples.} The denominator of $L_{GML}$ requires at least one contrast sample for each class. However, the strategy of MoCo~\cite{he2020momentum} does not guarantee any minimum number of samples, because it uses a queue of randomly sampled contrast samples. Therefore, we use multiple queues with different lengths. Each class has one assigned queue, and its length is proportional to the frequency of the class plus a predefined constant.
\begin{equation}
    \lVert Z_c \rVert=k_m+(k-k_m\times \lVert C \rVert)\times p(c)    
\end{equation}
where $k_m$ denotes the minimum length of the queue and $k$ denotes the total number of contrast samples.

\textbf{Teacher--Student Strategy.} Maintaining class-wise queues causes another problem. MoCo~\cite{he2020momentum} uses a momentum encoder to generate contrast samples, and they are stored in a queue. Therefore, old samples generated using an outdated encoder are replaced with new samples. On the other hand, we use multiple class-wise queues and their update frequency is proportional to the ratio of the classes in the training dataset. Therefore, queues of tail classes have excessively long update frequencies and old samples of their queues are generated by highly outdated encoders. To overcome this problem, we adopt a teacher--student strategy and use a pre-trained teacher encoder to generate contrast samples.

\textbf{Training Procedure.} Usually, a contrastive loss is used to train a backbone network, and a classifier layer is trained separately after training the backbone network. However, in the supervised setting, the classifier layer can be trained simultaneously with the backbone network. Therefore, we attach a cosine similarity classifier to the network and train them simultaneously. The classifier is trained without contrastive loss using the following loss function.
\begin{equation}
    -L_{cls}=\log\frac{\exp(m_x\cdot m_y/\tau_s+\eta_y )}
    {\sum_{c\in C}\exp(m_x\cdot m_c/\tau_s+\eta_c )}
\end{equation}
where $m_x$ denotes L2 normalized $x$, $m_c$ denotes L2 normalized weight at class $c$ of the classifier, and $\tau_s$ denotes a temperature hyperparameter for softmax cross-entropy loss. In contrast to the Gaussian mixture setting, $m_c$ is a parameter that needs to be trained.

In addition, we use MLP encoders followed by L2 normalization for contrast samples $z_x$ and $z_y$, similar to previous contrastive losses~\cite{chen2020simple,chen2020improved}, while a simple L2 normalization is used for the cosine similarity classifier.
\begin{equation}
    z_x=\frac{\text{MLP}(x)}{\lVert\text{MLP}(x)\rVert},\ m_x=\frac{x}{\lVert x \rVert}
\end{equation}

In summary, our training procedure is as follows. First, we estimate the ratio of classes in the training dataset to calculate $\eta_c$. Then, we train a teacher network with a cosine similarity classifier with $L_{cls}$ without contrastive loss. Finally, we train a student network and a cosine similarity classifier simultaneously with $L_{GML}$ and $L_{cls}$.

\textbf{Trainable Temperature.}
Unlike previous contrastive learning methods~\cite{chen2020simple,he2020momentum,khosla2020supervised,tian2020contrastive,cui2021parametric,zhu2022balanced}, we train $\tau_g$ along with the network parameters to reduce the burden of hyperparameter tuning. However, $L_{GML}$ is not suitable for training $\tau_g$. If we estimate the variance of the Gaussian mixture when the contrast centers include the query feature, the optimal variance is too small, and the Gaussian mixture becomes spiky. Therefore, we exclude contrast features extracted from the same sample to the query when training $\tau_g$. For simplicity, we use a fixed $\tau_g$ throughout most of the experiments in this work. However, we show in Sec.~\ref{sec:trainable_temperature} that $\tau_g$ can be trained, and it boosts the network performance.

By contrast, we find training $\tau_s$ results in a very low optimal value, which causes two negative effects. First, it makes $L_{cls}$ dominate the training procedure, and second, it significantly degrades the accuracy of tail classes. Therefore, we leave $\tau_s$ as a hyperparameter.

\subsection{Relation with Other Methods}~\label{sec:extending_gml}
As mentioned in Sec.~\ref{sec:proposed_formulation}, we can derive previous methods by adopting different likelihood and prior models.

\textbf{Balanced Softmax.} By modeling $s(x|y)$ as an isotropic Gaussian instead of a Gaussian mixture, we can derive the balanced softmax with a cosine similarity classifier.
\begin{equation}
    s(x|y)=\exp f(x,y)=\exp \left(\frac{-(m_x-m_y)^2}{2\sigma^2}\right)~\label{eq:isotropic_gaussian}
\end{equation}
where $m_y$ and $\sigma$ denote the center and variance of the Gaussian model, respectively.

Because we apply L2 normalization to $x$ and $y$, substituting $f$ in Eq.~\ref{eq:logit_compensate} results in $L_{cls}$, which is the cosine similarity classifier with logit adjustment.

\textbf{Supervised Contrastive Loss for Balanced Dataset.} Supervised contrastive loss~\cite{khosla2020supervised} can be derived by assuming the training dataset is balanced. A balanced training dataset gives all $\eta_c$s and $\lVert Z_c \rVert$s equal. Therefore, the proposed GML loss for balanced datasets becomes as follows:
\begin{align}
    -L_{GML}^{(balanced)}=&\log\frac{\sum_{z_y\in Z_y} \exp (z_x\cdot z_y /\tau_g)}{\sum_{c\in C}\sum_{z_c\in Z_c} \exp (z_x\cdot z_c /\tau_g)} \\
    \ge& \sum_{z_y\in Z_y} \log\frac{\exp (z_x\cdot z_y /\tau_g)}{\sum_{z\in Z} \exp (z_x\cdot z /\tau_g)} \nonumber
\end{align}
where $Z=\bigcup_c Z_c$ denotes the set of all contrast samples. By applying Jensen's inequality, we achieve supervised contrastive loss.

\textbf{Visual-Linguistic Representation Learning.} Visual-Linguistic Long-Tailed Recognition (VL-LTR)~\cite{tian2021vl} utilizes a text sentences dataset and a pre-trained visual-linguistic model to address the problem of insufficient training samples of tail classes. Similarly, our method can be used to connect visual and text representations. In particular, we use text features as centers of the Gaussian mixture instead of contrast features and the pre-trained visual-linguistic model to extract the text features.
\section{Experiments}
\subsection{Datasets}
\textbf{ImageNet-LT.} \citet{liu2019large} constructed ImageNet-LT by sampling ImageNet-2012~\cite{russakovsky2015imagenet} following a Pareto distribution with $\alpha=6$. The training set of ImageNet-LT contains $115.8k$ images of $1000$ classes, ranging from a maximum of $1280$ images to a minimum of $5$ images per class. Meanwhile, the test set is balanced such that head classes and tail classes have the same impact on the accuracy evaluation. The test set of ImageNet-LT contains $50k$ images, with $50$ images per class.

\textbf{iNaturalist 2018.} iNaturalist 2018~\cite{van2018inaturalist} is a large-scale image classification dataset containing $8142$ classes. The goal of iNaturalist 2018 is to push state-of-the-art image recognition for ``in the wild'' data of highly imbalanced and fine-grained categories. The training set of iNaturalist 2018 contains $437.5k$ images of $8142$ classes, ranging from a maximum of $1000$ images to a minimum of $2$ images per class. The test set of iNaturalist 2018 is also balanced, similar to ImageNet-LT.

\textbf{CIFAR-10-LT and CIFAR-100-LT.} \citet{cui2019class} constructed long-tailed versions of CIFAR~\cite{krizhevsky2009learning} by reducing the number of training samples according to an exponential function. CIFAR-LT is categorized by its imbalance factor, which is the ratio of training samples for the largest class to that for the smallest.

\textbf{ADE20K.} ADE20K~\cite{zhou2017scene} is a widely used image semantic segmentation dataset. The evaluation is conducted on $150$ classes, where the most common class comprises more than $15\%$ of total training pixels, while the rarest one comprises only $0.02\%$.

\subsection{Implementation Details}
We adopt training hyperparameter settings from previous long-tailed recognition papers~\cite{cui2021parametric,tian2021vl,zhu2022balanced} with some modifications. For ImageNet-LT, we train the proposed method using an SGD optimizer whose learning rate is initially set to $0.05$ and decays by a cosine scheduler. Input images are resized to $224\times224$, and a batch size of $128$ is used. The weight decay and momentum are set to $\num{5e-4}$ and $0.9$, respectively. The MLP of the contrast encoder has one hidden layer of $2048$ channels and its output layer has $1024$ channels. The total number of contrast samples ($k$) is $16384$, and the minimum number of contrast samples per class ($k_m$) is $2$. Randaugment~\cite{cubuk2020randaugment} is applied for the classifier training and Simaugment~\cite{chen2020simple} for the contrastive learning. Finally, $\tau_s=1/30$ is used throughout all experiments.

For iNaturalist 2018, we use a batch size of $128$, input image size of $224\times224$, and weight decay of $\num{2e-4}$. The learning rate is set to $0.02$ with the cosine scheduler. $k$ and $k_m$ are set to $65536$ and $2$, respectively. For CIFAR-LT, we use the training epochs of $200$ and $400$, a batch size of $64$, and a weight decay of $\num{5e-4}$. $k$ and $k_m$ are set to $4096$ and $2$, respectively. The learning rate is set to $0.05$ and decays by a factor of $10$ at $160$ and $180$ epochs ($320$ and $360$ epochs if the training epochs is $400$). We describe more about implementation details in Appendix~\ref{sec:more_implementation_details}.

\subsection{Long-Tailed Recognition}

\begin{table}[t]
\centering
\caption{Performance comparison on the ImageNet-LT dataset.}
\begin{adjustbox}{center}
\begin{tabular}{l|c|cccc}
\hline
\multirow{2}{*}{Method}               & \multirow{2}{*}{Epochs} & \multicolumn{4}{c}{Class frequency}                                                \\ \cline{3-6} 
                                      &                         & \multicolumn{1}{c|}{All}           & Many          & Med.          & Few           \\ \hline
Focal Loss        & 90                      & \multicolumn{1}{c|}{43.7}          & 64.3          & 37.1          & 8.2           \\
$\tau$-norm & 90                      & \multicolumn{1}{c|}{49.4}          & 59.1          & 46.9          & 30.7          \\
LWS         & 90                      & \multicolumn{1}{c|}{49.9}          & 60.2          & 47.2          & 30.3          \\
BALMS          & 90                      & \multicolumn{1}{c|}{51.4}          & 62.2          & 48.8          & 29.8          \\
LADE     & 90                      & \multicolumn{1}{c|}{51.9}          & 62.3          & 49.3          & 31.2          \\
DisAlign & 90                      & \multicolumn{1}{c|}{53.4}          & 62.7          & 52.1          & 31.4          \\
BCL            & 90                      & \multicolumn{1}{c|}{56.7}          & 67.2          & 53.9          & 36.5          \\
\textbf{Proposed}                     & \textbf{90}             & \multicolumn{1}{c|}{\textbf{58.3}} & \textbf{68.7} & \textbf{55.7} & \textbf{38.6} \\ \hline
PaCo         & 400                     & \multicolumn{1}{c|}{58.2}          & 68.0          & 56.4          & 37.2          \\
\textbf{Proposed}                     & \textbf{400}            & \multicolumn{1}{c|}{\textbf{58.8}} & \textbf{68.2} & \textbf{56.7} & \textbf{39.5} \\ \hline
\end{tabular}

\end{adjustbox}
\label{tab:imagenetlt}
\end{table}
\textbf{Comparison on ImageNet-LT.} We first compare the performance of the proposed method with existing state-of-the-art long-tailed recognition methods on the ImageNet-LT dataset. We compare performances of the same backbone ResNeXt-50 and same number of training epochs for a fair comparison. Following the previous categorization of classes~\cite{liu2019large}, we also evaluate the accuracy on subsets: many-shot (over 100 training samples), medium-shot (20-100 training samples), and few-shot (under 20 training samples).

\begin{table}[t]
\centering
\caption{Performance comparison on the iNaturalist 2018 dataset.}
\begin{tabular}{l|c|c}
\hline
Method                                  & Epochs       & Accuracy      \\ \hline
$\tau$-norm   & 100          & 65.6          \\
Hybrid-SC    & 100          & 66.7          \\
SSP           & 100          & 68.1          \\
KCL            & 100          & 68.6          \\
DisAlign   & 100          & 69.5          \\
RIDE (2 experts)    & 100          & 71.4          \\
BCL              & 100          & 71.8          \\
\textbf{Proposed}                       & \textbf{100} & \textbf{73.1} \\ \hline
RIDE (2 experts)    & 400          & 69.5          \\
$\tau$-norm   & 400          & 71.5          \\
Balanced Softmax & 400          & 71.8          \\
PaCo           & 400          & 73.2          \\
\textbf{Proposed}                       & \textbf{400} & \textbf{74.5} \\ \hline
\end{tabular}

\label{tab:inaturalist}
\end{table}
Table~\ref{tab:imagenetlt} presents the experimental results on the ImageNet-LT dataset. The proposed method shows the best overall performance, outperforming previous state-of-the-art methods significantly. The gain is maximized on tail classes proving the efficacy of the proposed method on the long-tailed recognition task.

\textbf{Comparison on iNaturalist 2018 dataset.} We also evaluate the performance of the proposed method on the iNaturalist 2018, which is a large-scale highly imbalanced image classification dataset. For a fair comparison with previous methods, we use ResNet-50 as backbone. Table~\ref{tab:inaturalist} shows the corresponding experimental result. The proposed method show significant performance improvements over all previous methods, including contrastive learning-based methods.

\begin{table}[t]
\centering
\caption{Performance comparison on the CIFAR-100-LT dataset with different imbalance factors.}
\begin{tabular}{l|c|ccc}
\hline
\multirow{2}{*}{Method}          & \multirow{2}{*}{Epochs} & \multicolumn{3}{c}{Imbalance factor}          \\ \cline{3-5} 
                                 &                         & 100           & 50            & 10            \\ \hline
Focal loss          & 200                     & 38.4          & 44.3          & 55.8          \\
CB-Focal            & 200                     & 39.6          & 45.2          & 58.0          \\
LDAM-DRW         & 200                     & 42.0          & 46.6          & 58.7          \\
BBN                  & 200                     & 42.6          & 47.0          & 59.1          \\
SSP           & 200                     & 43.4          & 47.1          & 58.9          \\
Casual model        & 200                     & 44.1          & 50.3          & 59.6          \\
Hybrid-SC    & 200                     & 46.7          & 51.8          & 63.1          \\
MetaSAug-LDAM     & 200                     & 48.0          & 52.3          & 61.3          \\
ResLT               & 200                     & 48.2          & 52.7          & 62.0          \\
BCL              & 200                     & 51.9          & 56.6          & 64.9          \\
\textbf{Proposed}                & \textbf{200}            & \textbf{53.0} & \textbf{57.6} & \textbf{65.7} \\ \hline
MiSLAS        & 400                     & 47.0          & 52.3          & 63.2          \\
Balanced Softmax & 400                     & 50.8          & 54.2          & 63.0          \\
PaCo           & 400                     & 52.0          & 56.0          & 64.2          \\
\textbf{Proposed}                & \textbf{400}            & \textbf{54.0} & \textbf{58.1} & \textbf{67.0} \\ \hline
\end{tabular}

\label{tab:cifar100lt}
\end{table}
\textbf{Comparison on CIFAR-100-LT dataset.} Subsequently, we conduct extensive experiments on the CIFAR-100-LT dataset with different imbalance factors. We adopt imbalance factors of $100$, $50$, and $10$, which are commonly used imbalance factors for evaluating the performance on CIFAR-LT dataset. A large imbalance factor implies a highly imbalanced dataset. In this experiment, we compare the performancess of ResNet-32 backbones.

Table~\ref{tab:cifar100lt} presents the comparison results on the CIFAR-100-LT dataset. The proposed method is robust to imbalance factors and consistently outperform previous long-tailed recognition methods on various imbalance factors by significant margins. Indeed, the robustness is also verified in experiments that compare our method with the supervised contrastive learning and knowledge distillation methods on balanced datasets. The comparisons are shown in Secs.~\ref{sec:exp_supcon} and \ref{sec:exp_kd}.

\textbf{Comparison with Visual-Linguistic Models.} Visual-linguistic models utilize training samples from text modality to enhance the performance of long-tailed image classification tasks. We compare our method with visual-linguistic models by extending it to learn the visual-linguistic representation as described in Section~\ref{sec:extending_gml}.
\begin{table}[t]
\centering
\caption{Performance comparison with visual-linguistic models. In this experiment, the backbone networks are initialized with CLIP~\cite{radford2021learning} pre-trained weights.}
\begin{tabular}{lcc}
\multicolumn{3}{l}{\textit{ImageNet-LT dataset:}}                                                                 \\ \hline
\multicolumn{1}{l|}{Method}                             & \multicolumn{1}{c|}{Backbone}           & Accuracy      \\ \hline
\multicolumn{1}{l|}{NCM}             & \multicolumn{1}{c|}{ResNet-50}          & 49.2          \\
\multicolumn{1}{l|}{cRT}             & \multicolumn{1}{c|}{ResNet-50}          & 50.8          \\
\multicolumn{1}{l|}{$\tau$-norm}     & \multicolumn{1}{c|}{ResNet-50}          & 51.2          \\
\multicolumn{1}{l|}{LWS}             & \multicolumn{1}{c|}{ResNet-50}          & 51.5          \\
\multicolumn{1}{l|}{Zero-Shot CLIP} & \multicolumn{1}{c|}{ResNet-50}          & 59.8          \\
\multicolumn{1}{l|}{VL-LTR}                  & \multicolumn{1}{c|}{ResNet-50}          & 70.1          \\
\multicolumn{1}{l|}{\textbf{Proposed}}                  & \multicolumn{1}{c|}{\textbf{ResNet-50}} & \textbf{70.9} \\ \hline
\multicolumn{1}{l|}{VL-LTR}                  & \multicolumn{1}{c|}{ViT-B}              & 77.2          \\
\multicolumn{1}{l|}{\textbf{Proposed}}                  & \multicolumn{1}{c|}{\textbf{ViT-B}}     & \textbf{78.0} \\ \hline
\\
\multicolumn{3}{l}{\textit{iNaturalist 2018 dataset:}}                                                            \\ \hline
\multicolumn{1}{l|}{Method}                             & \multicolumn{1}{c|}{Backbone}           & Accuracy      \\ \hline
\multicolumn{1}{l|}{VL-LTR}                  & \multicolumn{1}{c|}{ViT-B}              & 81.0          \\
\multicolumn{1}{l|}{\textbf{Proposed}}                  & \multicolumn{1}{c|}{\textbf{ViT-B}}     & \textbf{82.1} \\ \hline
\end{tabular}

\label{tab:vl_ltr}
\end{table}

Table~\ref{tab:vl_ltr} presents the comparison results with visual-linguistic models. We follow the training settings of VL-LTR~\cite{tian2021vl} and use a larger input size for the iNaturalist 2018 dataset. An input image size of $384\times384$ is used only for the iNaturalist 2018 dataset in this experiment. The proposed method successfully connects two different modalities, even when the dataset is imbalanced. Thus, it exhibits the best performance regardless of network architecture or dataset.

\subsection{Ablation Study}

\begin{table}[t]
\centering
\caption{Ablation study on the effect of each component.}
\begin{tabular}{c|c|c}
\hline
Loss type     & Teacher-student & Accuracy             \\ \hline
Cross-entropy & \xmark          & 55.4                 \\ \hline
BCL           & \xmark          & 56.7                 \\
BCL           & \cmark          & 57.1 (+0.4)          \\ \hline
Proposed      & \xmark          & 56.0                 \\
\textbf{Proposed}      & \cmark          & \textbf{58.3 (+2.3)} \\ \hline
\end{tabular}
\label{tab:ablation}
\end{table}

An ablation study is conducted on the ImageNet-LT dataset to investigate the effect of the components of the proposed method. Table~\ref{tab:ablation} reveals that there is a performance gain when the proposed loss is used. However, the gain is insufficient because the contrast samples of tail classes are generated by highly outdated encoders as described in Sec.~\ref{sec:training_gml}. Adopting a teacher--student framework solves the aforementioned problem, resulting in a significant gain in accuracy. To separate the effect of teacher--student framework from that of the proposed loss function, we measure the effect of teacher--student on BCL~\cite{zhu2022balanced}, which is another contrastive learning-based method for long-tailed recognition. Since BCL does not employ queues to store contrast samples, it does not suffer from the outdated encoder problem. The teacher--student framework does not significantly improve the performance on BCL, proving that the impact of knowledge distillation is not significant. Meanwhile, it resolves the outdated encoder problem of the proposed method and leads to significant performance improvement.

\subsection{Relation with Accuracy of Teacher} \label{sec:detailed_teacher}

To measure the effect of the accuracy of teacher and find the best one, we modulate the performance of teachers by adopting different sizes of backbone architecture. The chosen backbone architectures are ResNet-34, ResNet-50, ResNeXt-50, and ResNeXt-101, and they are trained on ImageNet-LT dataset for $90$ epochs. All students use the same backbone architecture ResNeXt-50 and are also trained for $90$ epochs. Table~\ref{tab:teacher} shows the effect of the performance of the teacher on the student. It is observed that the accuracy teacher dramatically decreases as the backbone is changed to ResNet-34, but the accuracy of the student remains stable. Moreover, the accuracy of the student surpasses that of the teacher. We find that a teacher with better performance leads to a better student, but the impact is marginal; a sufficient result can be achieved by using the same backbone architecture for both teacher and student. This result coincides with the finding of the ablation study, which indicates that the impact of knowledge distillation is not significant. Since a low-accuracy teacher can still successfully resolve the outdated encoder problem, the proposed method shows outstanding performance regardless teacher's accuracy.

\begin{table}[t]
\centering
\caption{Effect of Teacher's Performance on Student.}
\begin{tabular}{lc|lc}
\hline
\multicolumn{2}{c|}{Teacher}                & \multicolumn{2}{c}{Student}                \\ \hline
\multicolumn{1}{l|}{Backbone}    & Accuracy & \multicolumn{1}{l|}{Backbone}   & Accuracy \\ \hline
\multicolumn{1}{l|}{ResNet-34}   & 50.3    & \multicolumn{1}{l|}{ResNeXt-50} & 58.0    \\ \hline
\multicolumn{1}{l|}{ResNet-50}   & 55.2    & \multicolumn{1}{l|}{ResNeXt-50} & 58.1    \\ \hline
\multicolumn{1}{l|}{ResNeXt-50}  & 56.4    & \multicolumn{1}{l|}{ResNeXt-50} & 58.1    \\ \hline
\multicolumn{1}{l|}{ResNeXt-101} & 57.9    & \multicolumn{1}{l|}{ResNeXt-50} & 58.3    \\ \hline
\end{tabular}
\label{tab:teacher}
\end{table}

\subsection{Comparison with Supervised Contrastive Learning} \label{sec:exp_supcon}

\begin{table}[t]
\centering
\caption{Performance comparison with supervised contrastive learning on the CIFAR-10-LT dataset with different imbalance factors.}
\begin{tabular}{c|c|c|c}
\hline
Imb. factor & Cross-Entropy & SupCon & \textbf{Proposed} \\ \hline
1                & 94.8          & 96.0   & \textbf{95.9}     \\
10               & 88.4          & 94.0   & \textbf{94.5}     \\
50               & 69.1          & 88.1   & \textbf{90.6}     \\
100              & 64.1          & 82.7   & \textbf{86.7}     \\ \hline
\end{tabular}
\label{tab:comp_supcon}
\end{table}
Table~\ref{tab:comp_supcon} presents the performance comparison of the proposed method with supervised contrastive learning~\cite{khosla2020supervised}. Experiments are conducted using networks with ResNet-50 backbone on CIFAR-10-LT dataset with different imbalance factors. The proposed method shows the best accuracy with different imbalance factors, and the gap between previous methods increases as the imbalance factor increased. In addition, the proposed method shows a performance comparable with that of supervised contrastive learning when the dataset is balanced.

\subsection{Comparison with Knowledge Distillation} \label{sec:exp_kd}

\begin{table}[t]
\centering
\caption{Performance comparison with knowledge distillation methods.}
\begin{tabular}{lcc}
\multicolumn{3}{l}{\textit{Balanced dataset:}}                                                                  \\ \hline
\multicolumn{1}{l|}{Method}                         & \multicolumn{1}{c|}{CIFAR-100}     & ImageNet             \\ \hline
\multicolumn{1}{l|}{None}                           & \multicolumn{1}{c|}{69.1}          & 69.8                 \\
\multicolumn{1}{l|}{KD} & \multicolumn{1}{c|}{70.7}          & 70.7                 \\
\multicolumn{1}{l|}{CRD} & \multicolumn{1}{c|}{71.2}          & 71.2                 \\
\multicolumn{1}{l|}{\textbf{Proposed}}              & \multicolumn{1}{c|}{\textbf{71.4}} & \textbf{71.1}        \\ \hline
                                                    & \multicolumn{1}{l}{}               & \multicolumn{1}{l}{} \\
\multicolumn{3}{l}{\textit{Imbalanced dataset:}}                                                                \\ \hline
\multicolumn{1}{l|}{Method}                         & \multicolumn{1}{c|}{CIFAR-100-LT}  & ImageNet-LT          \\ \hline
\multicolumn{1}{l|}{None}                           & \multicolumn{1}{c|}{48.6}          & 56.3                 \\
\multicolumn{1}{l|}{KD} & \multicolumn{1}{c|}{49.9}          & 56.5                 \\
\multicolumn{1}{l|}{CRD} & \multicolumn{1}{c|}{50.6}          & 57.2                 \\
\multicolumn{1}{l|}{\textbf{Proposed}}              & \multicolumn{1}{c|}{\textbf{51.2}} & \textbf{58.3}        \\ \hline
\end{tabular}

\label{tab:comp_kd}
\end{table}

Because we adopt a teacher--student framework to train the proposed method, comparing it with previous knowledge distillation methods is relevant. We select two knowledge distillation methods for the comparison: KD~\cite{hinton2015distilling}, which does not utilize contrastive learning, and CRD~\cite{tian2020contrastive}, which utilizes contrastive learning. For the CIFAR-100 and CIFAR-100-LT datasets, we train a student network with ResNet-20 backbone using a teacher network with ResNet-56 backbone for $240$ epochs. We use a ResNet-18 student and a ResNet-34 teacher for ImageNet experiments, and a ResNeXt-50 student and a ResNeXt-101 teacher for ImageNet-LT experiments. For both ImageNet experiments, the networks are trained for $90$ epochs.

Table~\ref{tab:comp_kd} presents the performance comparison with knowledge distillation methods. The proposed method achieves the best performance on both balanced and imbalanced datasets. Knowledge distillation methods designed for balanced datasets show better accuracy than vanilla training on imbalanced datasets as well. This is because they provide measures to transfer the knowledge learned from head classes to tail classes, mitigating the lack of training samples. However, their gains are not the best because they do not consider the frequency of classes.

\subsection{Semantic Segmentation Task}

To prove the robustness of the proposed method, we replace the cross-entropy loss of semantic segmentation with the proposed method and measure the performance change. In this experiment, we measure the effect of the loss functions, not that of networks. Therefore, we perform the comparison using a widely used network FCN~\cite{long2015fully} with ResNet-50 backbone. The evaluation is conducted on ADE20K~\cite{zhou2017scene} using $160k$ training steps.

\begin{table}[t]
\centering
\caption{Performance comparison on the ADE20K semantic segmentation dataset.}
\begin{tabular}{l|c|c}
\hline
Method                  & mIoU           & mAcc           \\ \hline
Cross-Entropy           & 36.1          & 45.4          \\
\textbf{Proposed ($\boldsymbol{\alpha=0.2}$)} & \textbf{38.1} & 51.4         \\
\textbf{Proposed ($\boldsymbol{\alpha=1.0}$)} & 31.7          & \textbf{59.8} \\ \hline
\end{tabular}
\label{tab:ade20k}
\end{table}

Table~\ref{tab:ade20k} presents the performance comparison of losses. mIoU refers to the mean intersection-over-union (IoU) and mAcc refers to the mean accuracy (Acc), where the mean is taken over classes. The proposed method achieves the best mIoU or mAcc depending on the hyperparameter $\alpha$, which modulates the level of logit adjustment as given by Eq.~\ref{eq:alpha_gml_loss}.
\begin{align}
    &-L_{GML}^{(\alpha)}= \label{eq:alpha_gml_loss}\\
    &\log\frac{\exp\left[\log\frac{1}{\lVert Z_y \rVert}\sum_{z_y \in Z_y} \exp(z_x\cdot z_y/\tau_g)+\alpha\eta_y \right]}
    {\sum_{c\in C}\exp\left[\log\frac{1}{\lVert Z_c \rVert}\sum_{z_c \in Z_c} \exp(z_x\cdot z_c/\tau_g)+\alpha\eta_c \right]} \nonumber
\end{align}
\begin{align}
    \text{IoU}&=\text{TP}/(\text{TP}+\text{FN}+\text{FP}) \label{eq:def_iou}\\
    \text{Acc}&=\text{TP}/(\text{TP}+\text{FN}) \label{eq:def_acc}
\end{align}

\begin{figure}[t]
    \centering
    \includegraphics[width=\columnwidth]{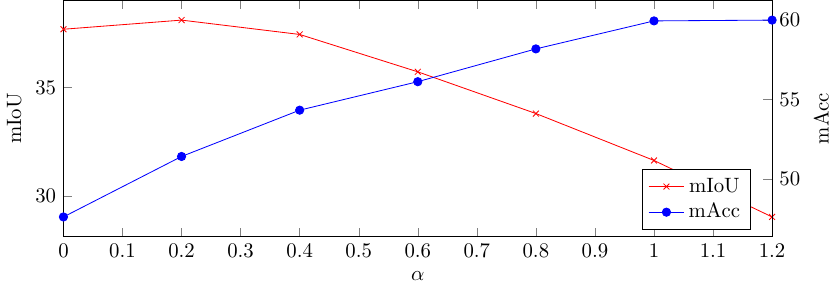}
    \vspace{-7mm}
    \caption{Effect of $\alpha$ on semantic segmentation performance.}
    \label{fig:semseg_alpha}
\end{figure}

Eqs.~\ref{eq:def_iou}-\ref{eq:def_acc} give the definitions of IoU and Acc, where TP, FN, and FP denote true positive, false negative, and false positive, respectively. mAcc is the same metric as the balanced evaluation setting used in classification tasks, and $\alpha=1.0$ gives the best mAcc. However, as FPs arise from other classes, mIoU is less sensitive to the accuracy of tail classes than mAcc. Therefore, the best mIoU is achieved by boosting the accuracy of head classes at the expense of tail classes, which is achieved by decreasing $\alpha$. The effect of $\alpha$ on mIoU and mAcc is shown in Fig.~\ref{fig:semseg_alpha}.

\subsection{Trainable Temperature Analysis} \label{sec:trainable_temperature}
\begin{figure}[t]
    \centering
    \includegraphics[width=\columnwidth]{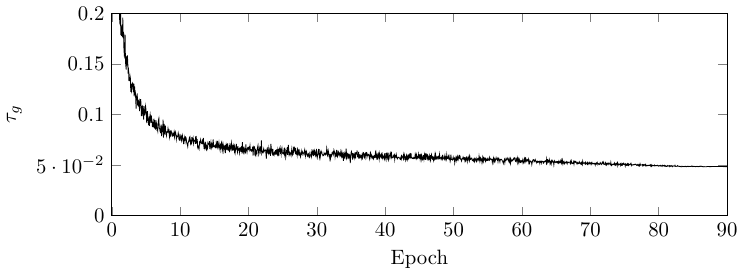}
    \vspace{-7mm}
    \caption{Change in $\tau_g$ during training.}
    \label{fig:tau_g}
\end{figure}

We examine the difference between a trainable $\tau_g$ and a fixed one. Fig.~\ref{fig:tau_g} shows the change in $\tau_g$ during training. $\tau_g$ becomes smaller as the encoder network converges. The final value is approximately $0.05$, which is similar to the hyperparameter choice of other methods~\cite{he2020momentum,tian2020contrastive,zhu2022balanced}, \ie, $0.07$. Furthermore, training $\tau_g$ results in slightly better performance, boosting the accuracy from $58.2$ to $58.3$.

\section{Conclusion}

In this paper, we show that the fundamental problem of contrastive learning methods on long-tailed recognition comes from maximizing the mutual information between latent features and input data. To overcome this limitation, we interpret the long-tailed recognition as the mutual information maximization between latent features and ground-truth labels. This approach seamlessly integrates contrastive learning and the logit adjustment technique. It also verifies that contrastive learning implies the use of a Gaussian mixture likelihood and the logit adjustment is derived from the prior, while previous methods have combined them without understanding the theoretical background. Further, we propose an efficient way of modeling the Gaussian mixture likelihood using a teacher--student framework.

Extensive experiments on both long-tailed datasets and balanced datasets verify the superiority of the proposed method, which marks state-of-the-art performance on various benchmarks. Finally, as real-world data often show a long-tailed distribution, the proposed method can be applied to other tasks as well. As an example, we conduct experiments on a semantic segmentation dataset. The proposed method also showed a large performance gain on semantic segmentation, demonstrating its versatility.
\section*{Acknowledgements}
This research was supported by the Challengeable Future Defense Technology Research and Development Program through the Agency For Defense Development(ADD) funded by the Defense Acquisition Program Administration(DAPA) in 2023(No.915027201), the Institute of New Media and Communications, the Institute of Engineering Research, and the Automation and Systems Research Institute at Seoul National University.

\bibliography{example_paper}
\bibliographystyle{icml2023}

\newpage
\appendix
\onecolumn

\setcounter{table}{0}
\renewcommand{\thetable}{A.\arabic{table}}
\setcounter{equation}{0}
\renewcommand{\theequation}{A.\arabic{equation}}

\section{Appendix}
\subsection{Contrastive Learning as MI Maximization between Latent Features and Input Data} \label{sec:detailed_contrastive_formulation}

The loss function of MoCo~\cite{he2020momentum} is written as follows.
\begin{equation}
    L_{MoCo}=-\log \frac{\exp(q\cdot k_+/\tau)}{\sum^K_{i=0}\exp(q\cdot k_i/\tau)} \label{eq:loss_moco_short}
\end{equation}
where $q$ denotes an encoded query, $k_i$ denotes $i$-th key, $k_+$ is a key in the dictionary that matches $q$, and $\tau$ is a temperature hyperparameter.

As MoCo uses separated data augmentations and encoders to extract query and keys, $q$ and $k$ can be rewrite as follows.
\begin{align}
    q&=f_q(t_q(x_j)) \label{eq:query_wo_param} \\
    k_+&=f_k(t_k(x_j)) \\
    k_i&=f_k(t_k(x_i))
\end{align}
where $f_q$ and $f_k$ denotes the query encoder and the key encoder, and $t_q$ and $t_k$ denotes data augmentations for query and key, respectively.

Finally, the back-propagation is blocked at the key encoder and only the query encoder is updated by gradient. To represent this, we update Eq.~\ref{eq:query_wo_param} to Eq.~\ref{eq:query_w_param}.
\begin{equation}
    q=f_q(t_q(x_j);w) \label{eq:query_w_param}
\end{equation}

Summarizing above, Eq.~\ref{eq:loss_moco_short} becomes as the follow.
\begin{equation}
    L_{MoCo}=-\log \frac{\exp(f_q(t_q(x_j);w)\cdot f_k(t_k(x_j))/\tau)}{\sum^K_{i=0}\exp(f_q(t_q(x_j);w)\cdot f_k(t_k(x_i))/\tau)} \label{eq:loss_moco_long}
\end{equation}

By substituting $x$, $y$, and $f$ in Eq.~\ref{eq:infonce} using the followings, we show that MoCo maximizes the mutual information between latent features and input data.
\begin{align}
    x_i &\leftarrow f_q(t_q(x_i);w) \\
    y_j &\leftarrow x_j \\
    f(x,y) &\leftarrow x\cdot f_k(t_k(y))/\tau
\end{align}

In other words, MoCo loss is identical to using a stochastic gradient descent to find the optimal parameter $w^*$ that maximizes a lower bound of mutual information between latent features, $f_q(t_q(x);w)$ and input data $x$.

\subsection{Implementation Details}
\label{sec:more_implementation_details}

Table~\ref{tab:implementation} shows the teacher architectures, training settings, augmentation strategies, and hyperparameter choices used in the experiments. $\gamma$, $\beta$, and $\alpha$ denote the weights used for $L_{cls}$, $L_{GML}$, and $L_{KD}$, respectively. We follow the settings of previous papers~\cite{cui2021parametric,zhu2022balanced} with some exceptions.

Training a network for $400$ epochs on ImageNet-LT leads to overfitting when $\gamma=1$ is used; reducing $\gamma$ while increasing $\beta$ alleviates the problem to improve the performance. Further, $L_{KD}$ considerably enhances the accuracy of CIFAR-LT experiments, but its effect becomes marginal when applied to ImageNet-LT or iNatualist experiments.

For experiments in Tables 4, 6, and 7, we follow the training settings and hyperparameter choices of baseline methods~\cite{khosla2020supervised,tian2020contrastive,tian2021vl}. In the experiments in Table 6, the encoder network and classifier are trained separately for a fair comparison with SupCon~\cite{khosla2020supervised}, whereas they are trained simultaneously in other experiments. The encoder network is trained for $1000$ epochs using $L_{GML}$. Subsequently, the classifier is trained for $100$ epochs using $L_{cls}$ with the encoder parameters fixed.

Semantic segmentation experiments in Table 8 are implemented based on open-source codebase mmsegmentation~\cite{mmseg2020} and follow the training hyperparameters and data augmentation settings provided in the codebase. The auxiliary loss of FCN~\cite{long2015fully} is replaced with the cosine similarity classifier and trained using $L_{cls}$. The segmentation head also is replaced with the cosine similarity classifier and trained using $L_{cls}+L_{GML}$. The hyperparameter choice is $(k,k_m,\tau_s,\tau_g)=(8192,27,0.05,0.07)$.

\begin{table}[t]
\caption{Hyperparameter choice of ImageNet-LT, iNaturalist, and CIFAR-100-LT experiments.}
\centering
\scriptsize
\begin{tabular}{c|c|c|c|c|c|c|c|c|c|c|c|c|c}
\hline
             & Arch\_s & Arch\_t & Epochs & MLP                & $k$     & $k_m$ & Aug\_cls & Aug\_GML & $\tau_g$ & $\tau_s$ & $\gamma$   & $\beta$ & $\alpha$ \\ \hline
Table 1. (a) & X50     & X101    & 90     & (2048, 2048, 1024) & 16384 & 2    & Randaug  & Simaug   & 0.07 & 1/30 & 1   & 1 & 0 \\ \hline
Table 1. (b) & X50     & X101    & 400    & (2048, 2048, 1024) & 16384 & 2    & Randaug  & Simaug   & 0.07 & 1/30 & 0.5 & 2 & 0 \\ \hline
Table 2. (a) & R50     & R152    & 100    & (2048, 2048, 1024) & 65536 & 2    & Simaug   & Simaug   & 0.1  & 1/30 & 1   & 1 & 0 \\ \hline
Table 2. (b) & R50     & R152    & 400    & (2048, 2048, 1024) & 65536 & 2    & Simaug   & Simaug   & 0.1  & 1/30 & 1   & 1 & 0 \\ \hline
Table 3. (a) & R32     & R56     & 200    & (64, 64, 32)       & 4096  & 2    & Autoaug  & Autoaug  & 0.1  & 1/30 & 1   & 1 & 1 \\ \hline
Table 3. (b) & R32     & R56     & 400    & (64, 64, 32)       & 4096  & 2    & Autoaug  & Autoaug  & 0.1  & 1/30 & 1   & 1 & 1 \\ \hline
\end{tabular}
\label{tab:implementation}
\end{table}


\end{document}